\documentclass[letterpaper, 10 pt, conference]{ieeeconf}  

\IEEEoverridecommandlockouts                              

\usepackage{amsmath} 
\usepackage{amssymb}  

\usepackage{cite}
\usepackage{algorithmic}
\usepackage{graphicx}
\usepackage{textcomp}
\usepackage{xcolor}
\usepackage{bm}
\usepackage{multirow}
\usepackage{siunitx}
\usepackage{tikz}
\usepackage{subcaption}
\graphicspath{{./fig/}}
\usetikzlibrary{arrows, calc,intersections}
\usetikzlibrary{arrows.meta}
\usetikzlibrary{decorations.pathmorphing}
\usetikzlibrary{decorations.text}
\usetikzlibrary{decorations.pathreplacing}
\usetikzlibrary{hobby,decorations.markings}
\usetikzlibrary{positioning}
\usetikzlibrary{calc}
\tikzset{%
	every neuron/.style={
		circle,
		draw,
		minimum size=0.15cm
	},
	neuron missing/.style={
		draw=none, 
		scale=1,
		text height=1,
		execute at begin node=\color{black}$\vdots$
	},
}

\usepackage[switch]{lineno}  
\usepackage[hyphens]{url}  

\def\BibTeX{{\rm B\kern-.05em{\sc i\kern-.025em b}\kern-.08em
		T\kern-.1667em\lower.7ex\hbox{E}\kern-.125emX}}

\usepackage{textcomp}
\newcommand\copyrighttext{%
	\footnotesize \textcopyright 2022 IEEE. Personal use of this material is permitted. Permission from IEEE must be obtained for all other uses, in any current or future media, including reprinting/republishing this material for advertising or promotional purposes, creating new collective works, for resale or redistribution to servers or lists, or reuse of any copyrighted component of this work in other works. DOI: 10.1109/ITSC55140.2022.9922419}
\newcommand\copyrightnotice{%
	\begin{tikzpicture}[remember picture,overlay]
	\node[anchor=south,yshift=10pt] at (current page.south) {\fbox{\parbox{\dimexpr\textwidth-\fboxsep-\fboxrule\relax}{\copyrighttext}}};
	\end{tikzpicture}%
}

\begin{document}
\title{\LARGE \bf
A Multidimensional Graph Fourier Transformation Neural Network\newline for Vehicle Trajectory Prediction}

\author{
	Marion Neumeier\textsuperscript{\rm 1}*\thanks{*We appreciate the funding of this work by AUDI AG.},
	Andreas Tollkühn\textsuperscript{\rm 3},
	Michael Botsch\textsuperscript{\rm 1} and 
	Wolfgang Utschick\textsuperscript{\rm 2}\\	
	\thanks{\textsuperscript{\rm 1} CARISSMA Institute of Automated Driving, Technische Hochschule Ingolstadt, 85049 Ingolstadt, Germany {\tt\small firstname.lastname@thi.de}}%
	\thanks{\textsuperscript{\rm 2} Technical University of Munich, 80333 Munich, Germany {\tt\small utschick@tum.de}}
	\thanks{\textsuperscript{\rm 3} AUDI AG, 85057 Ingolstadt, Germany {\tt\small andreas.tollkuehn@audi.de}}
}

\maketitle
\copyrightnotice


	\vspace{-10pt}
\begin{abstract}
	This work introduces the multidimensional Graph Fourier Transformation Neural Network (GFTNN) for long-term trajectory predictions on highways. Similar to Graph Neural Networks (GNNs), the GFTNN is a novel network architecture that operates on graph structures. While several GNNs lack discriminative power due to suboptimal aggregation schemes, the proposed model aggregates scenario properties through a powerful operation: the multidimensional Graph Fourier Transformation (GFT). The spatio-temporal vehicle interaction graph of a scenario is converted into a spectral scenario representation using the GFT. This beneficial representation is input to the prediction framework composed of a neural network and a descriptive decoder. Even though the proposed GFTNN does not include any recurrent element, it outperforms state-of-the-art models in the task of highway trajectory prediction. For experiments and evaluation, the publicly available datasets highD and NGSIM are used.
	%
	%
\end{abstract}

\section{Introduction}
Predicting the motion intention of nearby road users in different traffic scenarios is crucial for autonomous driving systems to operate safely. However, the trajectory prediction of other participants is a challenging task since it highly depends on the scenario's cooperative context. The cooperative context describes the situational and social interactions between traffic participants, that influence the behavior of each individual. These interdependencies of the cooperative context mainly result from the temporal and spatial relations between all participants. Existing approaches therefore attempt to model or statistically learn causalities within these two key dimensions. Since trajectory prediction can be regarded as a sequence-to-sequence learning task, most machine learning based methods use Recurrent Neural Networks (RNNs) \cite{Alahi.2016}\cite{Deo.2018}\cite{Messaoud.2021}.
Recently, Graph Neural Networks (GNNs) have also gained increasing popularity in the field of autonomous driving. GNNs are deep learning networks that operate on graph structures. Therefore, this class of deep learning architectures is especially powerful when applied on data generated from non-Euclidean domains or information represented by graphs. 

Although GNNs are successfully used in several applications \cite{Stokes.2020}\cite{DerrowPinion.25.08.2021}\cite{Carleo.2019}, there has been limited amount of works that study their representational properties.
Recent works with focus on mathematical properties revealed theoretical limitations of the representative power of GNNs \cite{Oono.020}\cite{Maron.2019} \cite{expressivenessGAT}. It has been shown, that popular GNN variants like the Graph Convolutional Network (GCN) \cite{GCN_KIPF} and GraphSAGE \cite{Hamilton.2017} lack expressive power, since these networks are unable to distinguish different graph structures \cite{Xu.01.10.2018}. The expressive power of any GNN mainly depends on the discriminative power of the aggregation scheme. A maximally powerful GNN provides an \textit{injective} neighborhood aggregation scheme, which means that two different neighborhoods are never mapped to the same representation \cite{Xu.01.10.2018}. GNNs that regard this principle are able to effectively learn complex relationships and interdependencies. 

In this work, the relational representation power of graphs is used to find an efficient scenario representation in order to predict a vehicle's trajectory. It introduces the multidimensional Graph Fourier Transformation Neural Network (GFTNN), which models the traffic scenario through a multidimensional graph. The aggregation of the scenario graph is done by applying the multidimensional Graph Fourier Transformation (GFT). This embedding is injective and preserves dimension properties. By forwarding the GFT embedding through a basic Feedforward Neural Network (FNN), the architecture outperforms current state-of-the-art deep neural networks in the task of trajectory prediction on highways. To the best of the authors' knowledge, this work is the first to simultaneously utilize spatial and temporal dependencies through the multidimensional GFT within a neural network setup for trajectory prediction in traffic. 

\textbf{Contribution.} The work contributes towards the usage of multidimensional dependencies for trajectory prediction through the definition of graphs and dimension perceiving operations. The main contributions are summarized as follows:
\begin{itemize}
	\item Introduction of the GFTNN for vehicle trajectory predictions on highways.
	\item Performance evaluation on the publicly available highway datasets highD and NGSIM.
	\item Interpretation of the prediction performance of the proposed GFTNN.
\end{itemize}
In this work, vectors are denoted as bold lowercase letters and matrices as bold capital letters.
\section{Related Work}
Being aware of the cooperative traffic context and the behaviour of individual traffic participants is essential for autonomous driving systems to safely plan their motion. Therefore, it is necessary to contrive architectures that are able to capture the interactive nature of traffic scenarios and predict the future motions of participants.
Motion prediction of vehicles can be categorized into two main categories: short-term predictions ($< \SI{1}{\second}$) and long-term predictions ($\geq \SI{1}{\second}$) \cite{Neumeier.2021}. Current state-of-the-art deep learning methods for long-term trajectory prediction of vehicles are mainly dominated by neural network architectures that are not based on graph structure.\cite{Mercat.2020}\cite{Messaoud.2021}\cite{Chaulwar.2017} An extensive survey on different approaches for motion prediction was done by Lefèvre \textit{et al.} \cite{Lefevre.2014}. Important representatives are the Social LSTM \cite{Alahi.2016} and Convolutional Social Pooling LSTM (CS-LSTM) \cite{Deo.2018}. These models use RNNs and Convolution Neural Networks (CNNs) to evaluate the sequential and interdependent nature of the trajectory prediction task. Messaoud \textit{et al.} \cite{Messaoud.2021} extended this idea by introducing the Multi-head Attention Social Pooling (MHA-LSTM) model. The core idea of the MHA-LSTM is to use the attention mechanism \cite{Vaswani.2017} to better evaluate the cooperative context of a scenario. Although such RNN-based models are widespread and perform good in sequence analysis, these methods suffer from time-consuming iterative propagation and gradient explosion/vanishing issues \cite{Pascanu.21.11.2012}. The proposed GFTNN does not contain recurrent network structures for the prediction and therefore prevents these issues.

Over the last years, the field of GNNs has experienced increasing attention and various models were introduced. For spatio-temporal sequence learning tasks, which relates to the trajectory prediction task, often spatio-temporal Graph Neural Networks (STGNN) are applied. The surveys \cite{Zhou.2020} and \cite{Wu.2021} provide an introduction to the group of STGNN. In \cite{Yu.2018}, a STGNN was applied on traffic forecasting. The STGNN model is based on a graph structure followed by operations of stacked sequential convolutions. In contrast to the proposed network in this work, their algorithm does not define a spatio-temporal graph but rather creates a stacked set of spatial graphs. In \cite{Mo.08.07.2021}, a graph and RNN based vehicle trajectory prediction model for highway driving was introduced. The model includes a Time-Extrapolator Convolution Neural Network (TXPCNN) layer to setup a stateless system for the trajectory prediction.
Zhou \textit{et al.} \cite{Zhou.2021} propose an Attention-based Spatio-Temporal Graph Neural Network (AST-GNN) for interaction-aware pedestrian trajectory prediction. This model uses the attention mechanism in order to extract interactions and motion patterns within the spatial-temporal domain. Contrary to the architecture proposed in this work, the AST-GNN handles the spatio-temporal dimensions separately. It has a spatial GNN and an additional temporal GNN. 
In \cite{RAGAT.2021}, the Repulsion and Attraction Graph Attention (RA-GAT) model for trajectory prediction is presented. The model is based on two stacked Graph Attention Networks \cite{GAT.2018}, that address either free space or vehicle state information through distinct graph definition. Through this setup, the authors follow the idea of repulsive and attractive forces within a traffic scenario. To encode and decode the vehicle movements, LSTMs are used.
VectorNet, introduced in \cite{Gao.08.05.2020}, is a hierarchical GNN that initially aggregates the agents’ trajectories and map features to polyline subgraphs. The information is then passed to a global interaction graph (GNN) to fuse the features among the subgraphs. The global graph aggregation function is implemented through a self-attention mechanism. 

The approach most related to the one of this work is the Spectral Temporal Graph Neural Network (SpecTGNN) of Cao \textit{et al.} \cite{Cao.5302021652021}. The SpecTGNN handles the environmental and agent modelling separately. Both matters are computed by defining a 1D spatial connectivity graph structure, where each node contains a feature vector stacking the information of all considered time steps. By separately applying a spectral graph convolution and a temporal convolution, the state information of the interactive agents (and the environment) are encoded. After combining the resulting spatial and temporal latent spaces, the inverse Fourier transform is applied. Following, the environmental and agent modelling latent spaces are added and a multi-head attention mechanism is applied. The trajectory is predicted through a temporal CNN. In contrast to the proposed model of this work, the SpecTGNN does not define a spatio-temporal graph, but rather addresses both dimensions separately and successively. 
\section{Method}
The proposed GFTNN is novel due to the extraction of expressive representations of traffic scenarios through a global and injective aggregation function based on the graph properties. 
By defining a spatio-temporal graph based on the scenario and applying the multidimensional GFT, the graph's spectral features are computed. This spectrum holds the temporal and spatial dynamic characteristics of a scenario. The transformed scenario representation is input to a neural network. The GFTNN does not contain any computationally expensive components for the trajectory prediction and thus model complexity is reduced.
\subsection{Preliminaries}
\textbf{Graph definition}: Let $\mathcal{G} = (\mathcal{V},\mathcal{E}, \bm{W})$ be an undirected weighted graph with $\mathcal{V}$ being the set of nodes, $\mathcal{E}$ the set of edges and the weight matrix $\bm{W} \in \mathbb{R}^{N \times N}$. The weight matrix is symmetric and satisfies $\bm{W}_{ij} = 1$ for all connections $i \rightarrow j$ if the graph is unweighted. The graph consists of $N=|\mathcal{V}|$ nodes and $E = |\mathcal{E}|$ edges. The connectivity of the graph is represented by the adjacency matrix $\bm{A} \in \mathbb{R}^{N\times N}$ where $\bm{A}_{ij} = 1$ if there exists an edge between the nodes $i$ and $j$, otherwise $\bm{A}_{ij} = 0$. The degree matrix $\bm{D} \in \mathbb{R}^{N\times N}$ is a diagonal matrix with the degrees on the diagonals such that $\bm{D}_{ii} = \sum_{j} \bm{W}_{ij}$. The node feature of node $i$ is denoted by $f(i)$. Particularly important for the GFT is the Laplacian matrix $\bm{L} = \bm{D} - \bm{W} \odot \bm{A}$, where $\odot$ indicates the Hadamard product operation. The Laplacian matrix $\bm{L} \in \mathbb{R}^{N\times N}$ is a real, symmetric and positive-semidefinite matrix.
\begin{figure}
		\vspace{9pt}
	\centering
	\includegraphics[width=0.98\columnwidth]{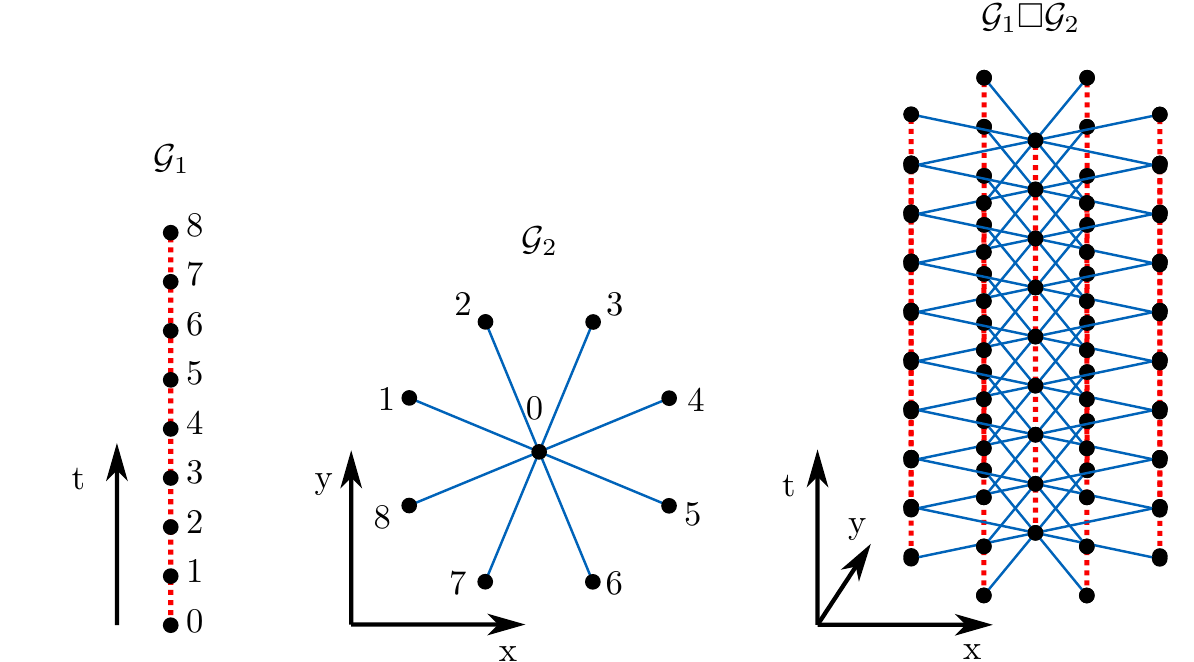}
	\caption{Example graphs $\mathcal{G}_1$ and $\mathcal{G}_2$ and their Cartesian product graph $\mathcal{G}_1\square \mathcal{G}_2$. Red dotted edges in $\mathcal{G}_1 \square  \mathcal{G}_2$ result from graph $\mathcal{G}_1$ and blue solid edges come from $\mathcal{G}_2$ \cite{Kurokawa.2017}.}
	\label{fig:CartesianProductGraph}
\end{figure}

\textbf{Cartesian product graph}:
A Cartesian product $\mathcal{G}_1\square \mathcal{G}_2$ is a type of graph multiplication that satisfies specific properties as described in \cite{Kurokawa.2017}. Such a Cartesian product operation for the graphs $\mathcal{G}_1$ and $\mathcal{G}_2$ is qualitatively illustrated in Fig.~\ref{fig:CartesianProductGraph}. A beneficial property of the Cartesian product graph is that solving the eigenproblem of the resulting graph $\mathcal{G}_1 \square \mathcal{G}_2$ can be broken down into addressing the eigenproblem of each factor graph $\mathcal{G}_1$ and $\mathcal{G}_2$ separately.

\textbf{Eigenvalue decomposition of graphs}: The eigenvalue problem of graphs can be solved by using the Laplacian matrix $\bm{L}$ \cite{Kurokawa.2017}. Due to the mathematical properties of the Laplacian matrix $\bm{L} \in \mathbb{R}^{N\times N}$ its eigenvalue decomposition results in $N$ real, non-negative eigenvalues $\lambda_0,\dots, \lambda_{N-1}$ and the eigenvectors $\bm{u}_0,\dots, \bm{u}_{N-1}$ such that
\begin{equation}
\bm{L}\bm{u}_i = \lambda_i\bm{u}_i,
\label{eq:EIG}
\end{equation}
where index $i = 0, \dots, N-1$. The resulting set of eigenvalues (usually sorted by size) are denoted as the spectrum of a graph. The corresponding eigenvectors form the orthogonal basis $\mathfrak{B}=\{\bm{u}_0,\dots, \bm{u}_{N-1}\}$ which can be used to perform frequency transformations. 

\textbf{Multidimensional graph Fourier transformation}:
Analogous to the classic Fourier transformation, the set of eigenvalues $\boldsymbol{\Lambda} \in \mathbb{R}^N$ of a Laplacian matrix represent frequencies and the eigenvectors $\bm{U} \in \mathbb{R}^{N \times N}$ form the basis $\mathfrak{B}$. 
The multidimensional GFT of a Cartesian product graph $\mathcal{G}_1 \square \mathcal{G}_2$ of the signal $f$ is described by \cite{Kurokawa.2017}
\begin{equation}
\hat{f}(\lambda_{l_1}^{(1)}, \lambda_{l_2}^{(2)}) =
\sum^{N_1-1}_{i_1=0} \sum^{N_2-1}_{i_2=0}
f(i_1, i_2) \overline{u_{l_1}^{(1)}(i_1)u_{l_2}^{(2)}(i_2)},
\label{eq:GFT}
\end{equation}
for  $l_1 = 0, \dots, N_1-1$ and  $l_2 = 0, \dots, N_2-1$. The superscript $^{(1)}$ denotes the eigenvalues/eigenvectors of graph $\mathcal{G}_1$ and $^{(2)}$ of graph $\mathcal{G}_2$, respectively. The expression $\overline{u_{l_1}^{(1)}(i_1)u_{l_2}^{(2)}(i_2)}$ denotes the element-wise complex conjugate of the multiplication. 

\subsection{Problem Definition}
The task is to predict the trajectory of a single target vehicle in a traffic scenario, given motion observations of the target vehicle and its surrounding vehicles. In total, the $N_V$ road users including the target vehicle itself are considered in the prediction process. At each timestep $t$ the relative spatial coordinates of the surrounding road users with respect to the target vehicle are given. The motion information of the $ j $-th participating road user is represented as ${\boldsymbol{\xi}_j= [\mathbf{x}_{j,\textrm{rel}}, \mathbf{y}_{j,\textrm{rel}}, \mathbf{v}_{j,x,\textrm{rel}}, \mathbf{v}_{j,y,\textrm{rel}}]}$. The input matrix of the $j$-th road user contains the relative distances ($\mathbf{x}_{j,\textrm{rel}}, \mathbf{y}_{j,\textrm{rel}}$) and velocities ($\mathbf{v}_{j,x,\textrm{rel}}, \mathbf{v}_{j,y,\textrm{rel}}$) within a specified observation period $ t_\textrm{obs}$ up to the
current time step $ t_0 $. The scenario origin is fixed within the start position of the target vehicle. The input of the prediction module is given through ${\bm{X} = [\boldsymbol{\xi}_1, \boldsymbol{\xi}_2, \dots , \boldsymbol{\xi}_{N_V}]^\textrm{T}}$ where $\boldsymbol{\xi}_1$ is the prediction vehicle's motion information. By processing the input relations, the model predicts the trajectory of the target vehicle $ {\mathbf{\hat{c}} = [\mathbf{\hat{x}}^\textrm{T}, \mathbf{\hat{y}}^\textrm{T}]^\textrm{T} }$ within the prediction period from $ t_0 $ to $ t_\textrm{pred}$. The total number of time steps covered by the observation and prediction period are given by $T_{\text{obs}}$ and $T_{\text{pred}}$, respectively.

\subsection{Model Structure}
\begin{figure*}[ht]
	\centering
	\vspace{4pt}
	\includegraphics{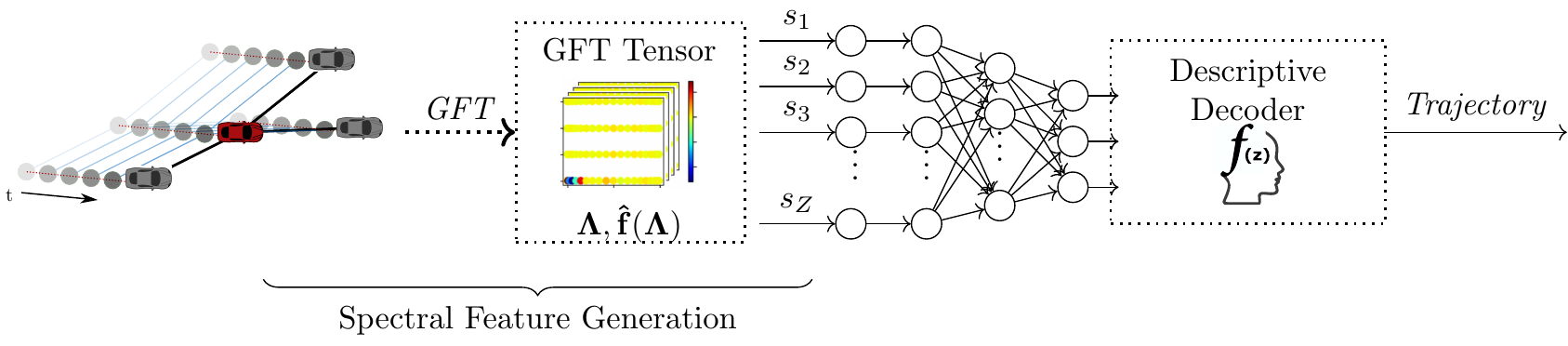}
	\caption{Architecture of the GFTNN. The observed traffic scenario is represented as a 2D graph and mapped into a frequency scenario representation through the GFT. The GFT tensor operates as the input to a MLP that computes the parameters of the descriptive decoder in order to predict the future trajectory of the target vehicle.}
	\label{fig:gilbert_architecture}
\end{figure*}
The model is structured into three elements, namely the spectral feature generation, an encoder and a decoder. In Fig.~\ref{fig:gilbert_architecture}, the complete architecture of the prediction model resulting from the sub-elements is shown. The proposed GFTNN represents a scenario as 2D graph. From this graph representation, spectral features are extracted by applying the multidimensional GFT and these features are used as input to a Multilayer Perceptron (MLP).

In a first step, a multidimensional graph structure is built through the Cartesian product graph considering the temporal and spatial interdependencies. The definition of graph $\mathcal{G}_\mathrm{S}$ is based on the spatial relation between the traffic participants of a scenario. Within the graph, each node represents an agent and the edges describe possible interdependencies due to cooperative interactions. The spatial graph construction is defined based on the interaction assumptions between the traffic participants, e.\,g.: 
\begin{itemize}
	\item \textit{Spider graph}: All traffic participants primarily interact with the target vehicle, but not among each other,
	\item \textit{Mesh graph}: All traffic participants interact with the target vehicle and also among each other, and
	\item \textit{Scenario-dependent graph}: Dependent on pre-defined rules that define interactions, the graph can be build up accordingly, e.\,g., $k$-Nearest-Neighbor or $\epsilon$-Neighborhoods.
\end{itemize}

In the proposed GFTNN, a spider graph is applied and hence, the spatial graph definition is similar to graph $\mathcal{G}_2$ in Fig.~\ref{fig:CartesianProductGraph}. Note, that there also might be interdependencies between neighboring agents. Despite not modelling these explicitly through the spider graph, these interdependecies are implicitly transformed into the spectral scenario since the prediction vehicle itself is connected to all participants. If the goal was a multi-modal prediction, it would be reasonable to explicitly model all connections by using a mesh graph. 
To represent the temporal dependencies of each vehicle, a line-graph is set up. Each vehicle has dependencies with itself as time evolves. Therefore, each step of the observed time sequence represents a node and each node is connected to its previous and following time step node.
The resulting temporal graph $\mathcal{G}_\mathrm{T}$ is qualitatively similar to graph $\mathcal{G}_1$ of Fig.~\ref{fig:CartesianProductGraph}. Hence, the resulting Cartesian product graph of $\mathcal{G}_\mathrm{T} \square \mathcal{G}_\mathrm{S} \cong \mathcal{G}_1 \square \mathcal{G}_2$.

In the proposed GFTNN, the graph definition is kept constant and is not adopted depending on the scenario. The multidimensional GFT can be considered as the aggregation scheme of the graph structure. As stated initially, such graph aggregations are optimally chosen to be injective. Since the GFT is based on the eigendecomposition of the graph, which is a non-injective operation, also the aggregation would be non-injective if the graph was non-static. By keeping the graph definition constant the GFT represents an injective aggregation function.
In appendix A, a proof for the injective nature of the GFT based on a static graph is provided.
\subsection{Spectral Feature Generation}
\begin{figure}
		\vspace{9pt}
		\begin{subfigure}{0.49\columnwidth}
			\centering
			\tiny
			\def\svgwidth{\columnwidth}
			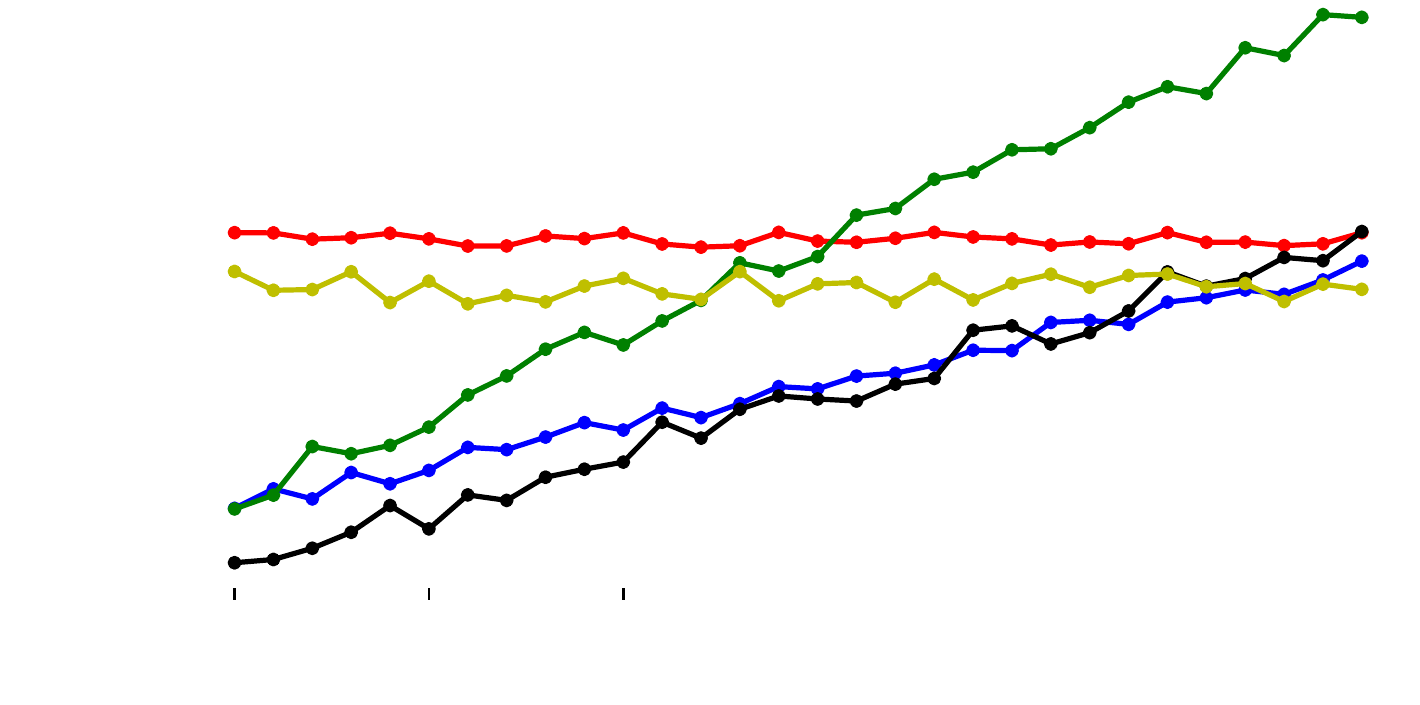
			\caption{ }
			\label{subfig:temp}
		\end{subfigure}
		\begin{subfigure}{0.49\columnwidth}
			\centering
			\tiny
			\def\svgwidth{\columnwidth}
			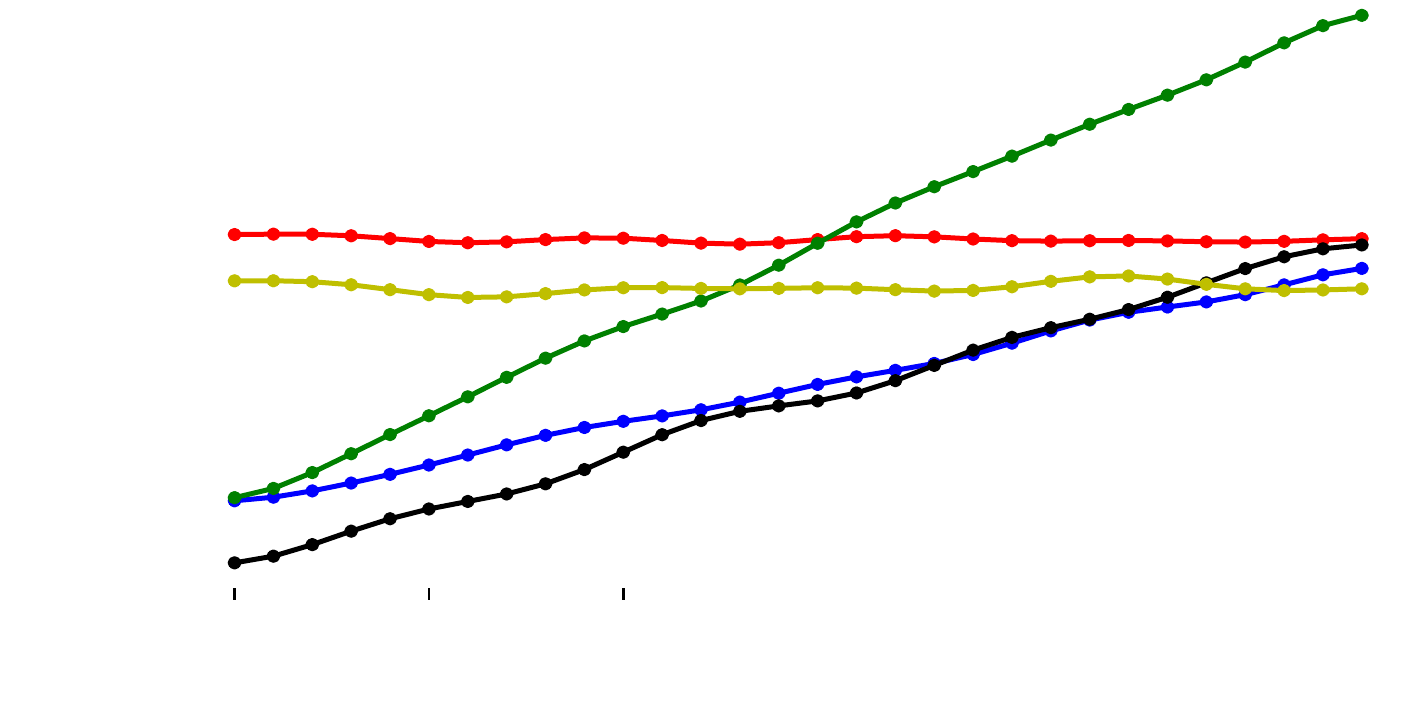
			\caption{ }
			\label{subfig:inv}
		\end{subfigure}
		\hspace{1 cm}
		\begin{subfigure}{\columnwidth}
			\centering
			\def\svgwidth{\columnwidth}
			\scriptsize
			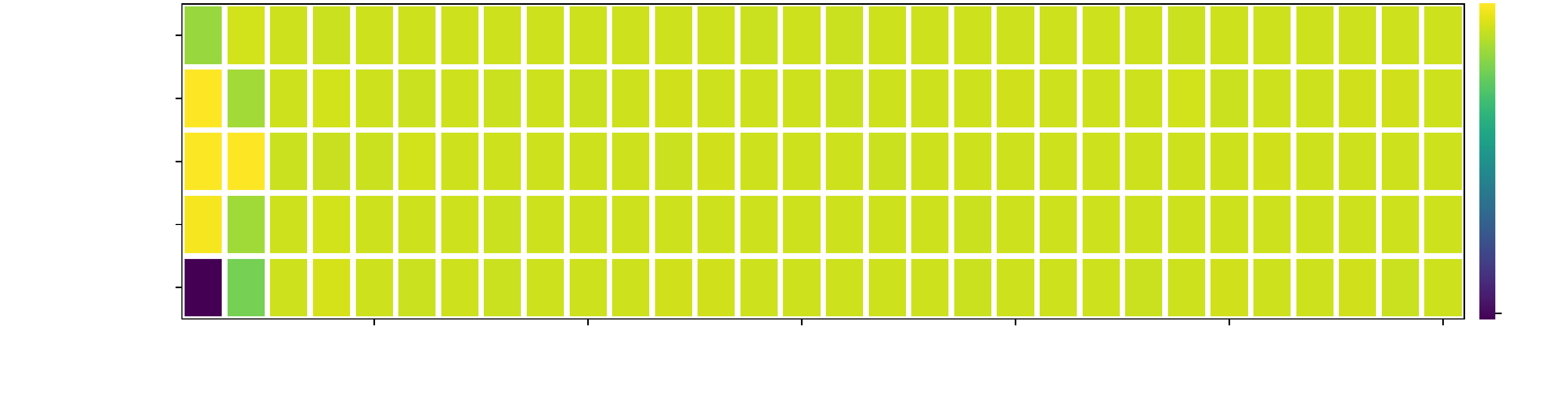
			\caption{ }
			\label{subfig:spectrum}
		\end{subfigure}
		\caption{The traffic scenario regarding the feature velocity in (a) is mapped into a frequency scenario representation (c) through a multidimensional GFT. By applying the inverse GFT that only considers the $p$ smallest eigenvectors results in a low-pass filtered time signal as shown in (b).}
	\end{figure}

The spectral feature generation of the GFTNN is similar to a deterministic aggregation scheme within GNNs. 
Along with the graph definition, the Laplacian matrices $\bm{L}_\mathrm{T} \in \mathbb{R}^{T_{\text{obs}}\times T_{\text{obs}}}$ and $\bm{L}_\mathrm{S} \in \mathbb{R}^{N_V \times N_V}$ are computed, where $T_{\text{obs}}$ is the dimension of the observation time steps and $N_V$ the total number of traffic participants in a scenario.
The eigenvalues $\boldsymbol{\Lambda}^{(\mathrm{T})}, \boldsymbol{ \Lambda}^{(\mathrm{S})}$ and eigenvectors $\bm{U}^{(\mathrm{T})}, \bm{U}^{(\mathrm{S})}$ result from the individual eigenvalue decomposition (c.\,f. (\ref{eq:EIG})) of the Laplacian matrices $\bm{L}_\mathrm{T}$ and $\bm{L}_\mathrm{S}$. Based on this information, the scenario graph is transformed into the spectral domain by a multidimensional GFT. The advantage of the multidimensional GFT is that the transformation retains the dimensional context of the graph signal. This differentiation of the spatial and temporal domain within the graph spectrum would be lost by using only an one-dimensional GFT. Since the dynamics along these two dimensions imply different scenario conditions, it is meaningful to keep the multidimensional nature. Semantically, the spectral feature space indicates how smooth a vehicle's movement is with respect to the neighboring vehicles and along time.
Without loss of generality, the signal function is extended in a fashion that each node of the graph holds a feature vector information such that $\bm{F} \in \mathbb{R}^{K \times N_1 \times N_2}$. Features can be for example: velocity, acceleration, etc. Totally there are $K$ features. The extended formula describing the GFT, considering the extended feature information, is
	\begin{equation}
	\bm{\hat{F}}(k, \lambda_{l_1}^{(1)}, \lambda_{l_2}^{(2)}) = 
	\sum^{N_1-1}_{i_1=0} \sum^{N_2-1}_{i_2=0}
	\bm{F}(k, i_1, i_2) \overline{u_{l_1}^{(1)}(i_1)u_{l_2}^{(2)}(i_2)},
	\label{eq:extendedGFT}
	\end{equation}
	where $k = 1, \dots, K$ represents the feature dimensions.
	The spectral representation maps the temporal and spatial dynamic within a traffic scenario under consideration of the context interdependencies. The transformation results in a GFT tensor $\bm{\hat{F}} \in \mathbb{R}^{K \times N_1 \times N_2}$ where $N_1 \equiv N_{T_{\text{obs}}}$ is the total amount of observed time steps and $N_2 \equiv N_V$ the considered traffic participants including the target vehicle. 
	An exemplar of a spectral scenario representation for a one-dimensional feature information with $K=1$ can be seen in Fig.~\ref{subfig:spectrum}. The spectral representation results from the GFT of the scenario illustrated in Fig.~\ref{subfig:temp}, where the velocity information for 5 different vehicles is given over a period of 30 time steps.
	The information of the resulting spectrum can be interpreted as classical frequencies: Small (eigen-)values indicate low frequencies and vice versa. For this application, the high frequencies can be interpreted as noise, hence, they do not carry important scenario information. Large eigenvalues within the temporal spectra can possibly be neglected for further computations, since the important scenario characteristic is represented through the small eigenvalues. By selecting the $p$ most important eigenvalues $\boldsymbol{\Lambda}^{(\mathrm{T})}_{[0:p-1]}$ irrelevant information for the trajectory prediction can be filtered out. The filtering process within the frequency domain results in a low-pass characteristics in the original time domain. The resulting low-pass characteristic can be seen in Fig.~\ref{subfig:inv}, where the inverse GFT is applied by only using the $p=10$ smallest eigenvalues $\boldsymbol{\Lambda}^{(\textrm{T})}_{[0:p-1]}$ of the spectrum illustrated in Fig.~\ref{subfig:spectrum}. Since no inverse GFT is applied within the GFTNN, the hyper-parameter $p$ defines the temporal eigenvalue information used for further computations. Hence, from the GFT representation $\bm{\hat{F}}$ the constant subsection of $\bm{s} = \left[s_1, s_2, \dots, s_{Z}\right]$, where $Z = |{K\times N_{T_{\text{obs}}[0:p-1]}\times N_V}|$ is chosen to be forwarded as input to the encoder.
	
	\subsection{Encoder}
\begin{figure}[t!]
		\begin{center}
			\vspace{9pt}
			\includegraphics{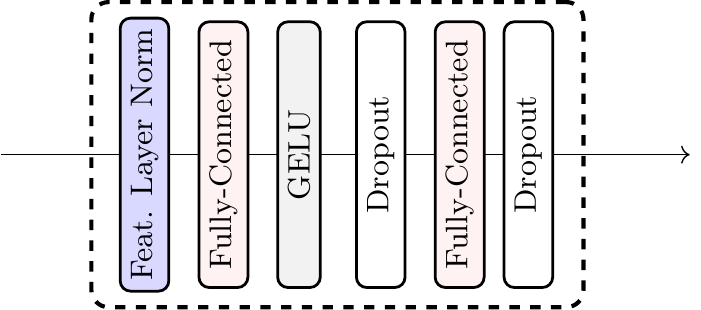}
			\caption{Feature-selective MLP block architecture.}
			\label{fig:encoder}
		\end{center}
	\end{figure}

	The encoder is designed as lightweight neural network which does not contain any recurrence or computationally expensive component. Its core element is the MLP block as illustrated in Fig.~\ref{fig:encoder}. After an element-wise multiplication of the GFT scenario representation 
	\begin{equation}
	\bm{h}_s = \bm{s} \odot \bm{w}_s,
	\end{equation}
	where $\bm{w}_s \in \mathbb{R}^Z$ are learnable parameters, the result is passed on to the MLP block. This block is stacked and computationally repeated $N_x$ times. Note, that each of the $K$ features is passed through a separate MLP block. The feature-selective information $\bm{h}^k_{l}$ are concatenated afterwards. Mathematically, the MLP block is represented as
	\begin{align}
	\bm{h}^k_{\text{norm}} &= \frac{\bm{h}_s^k - \mathbb{E} [\bm{h}_s^k]}{\sqrt{\text{VAR}[\bm{h}_s^k] + \epsilon}}, \quad \epsilon = 1e-05 \label{eq:layernorm}\\
	\bm{h}^k_{l} &= \bm{W}^k_{l}\,\Phi(\bm{W}^k_{n} \bm{h}^k_{\text{norm}} + \bm{b}^k_n) + \bm{b}^k_l\\
	\bm{h}_c &= concat(\bm{h}^0_{l}, \dots, \bm{h}^K_{l}),
	\end{align}
	where $\bm{h}_s^k \in \mathbb{R}^{|N_{T_{\text{obs}}[0:p-1]}\times N_V|}$ is the corresponding feature information from $\bm{h}_s$ and $\Phi$ is the non-linear activation through the GELU function \cite{Hendrycks.27.06.2016}.
	Following \cite{ba2016layer}, the layer normalization (\ref{eq:layernorm}) computes the mean $\mathbb{E}[\cdot]$ and variance $\text{VAR}[\cdot]$ based on a single feature dimension and training case.
	The combined information vector $\bm{h}_c$ is forwarded through a sigmoid function and another fully connected linear layer, that outputs the latent space representation $\bm{h}_z \in \mathbb{R}^3$. 
	\subsection{Decoder}
	The decoder of the GFTNN is implemented as the descriptive decoder proposed in \cite{Neumeier.2021}. Instead of using neural networks for creating the future trajectory, a model-based approach is used. The usage of the descriptive decoder setup introduces interpretability in the latent space $\bm{h}_z$.
	In this descriptive decoder the longitudinal ($\hat{\bm{x}}$) and lateral ($\hat{\bm{y}}$) trajectory are approximated through functions capable of representing the vehicle dynamics on highways. This simple vehicle dynamic model provides three parametrizable variables, that define the trajectory characteristic. The parametrization is based on the computations of the encoder network: The variables of the latent space $\bm{h}_z$ are used as variables in the model-based decoder where
	\begin{equation}
	\bm{\hat{x}} = v_0\bm{t} + 0.5h_{z_1}\bm{t}^2,
	\end{equation}
	\begin{equation}
	\bm{\hat{y}} = \frac{h_{z_2}}{1+e^{h_{z_3}\boldsymbol{ \tau}}} - \frac{h_{z_2}}{1+e^{h_{z_3}{ \tau_0}}} .
	\end{equation}
	The variables $\bm{t} = 0, \dots, T_{pred}$ and $\boldsymbol{ \tau} = \bm{t} - 0.5T_{\text{pred}}$ define the temporal information of the trajectory, and $v_0$ is the latest observed longitudinal velocity of the prediction vehicle. 
	\begin{table*}[t!]
		\centering
		\vspace{7pt}
		\begin{tabular}{c|c||c|c|c|c|}
			\multirow{2}{*}{Architecture}&
			\multirow{2}{*}{$n_{params}$}&
			\multicolumn{2}{c|}{highD} & 
			\multicolumn{2}{c|}{NGSIM} \\
			&&ADE[\SI{}{\meter}]@\SI{5}{\second} 
			& FDE[\SI{}{\meter}]@\SI{5}{\second} & ADE[\SI{}{\meter}]@\SI{5}{\second} & FDE[\SI{}{\meter}]@\SI{5}{\second}  \\
			\hline
			\hline
			\textbf{gftnn}	    & 567.253 	&\textbf{1.04} & \textbf{2.41} & \textbf{3.08} & 7.31  \\	\hline
			\textbf{gftnn}-w       & 283.653 	& 1.15 & 2.60 & 3.13 & \textbf{7.21} \\ 	\hline
			\textbf{gftnn}-rdcby5 & 113.653 	& 1.10 & 2.58 & 3.56 & 8.75 \\	\hline
			\textbf{gftnn}-rdcby15& 38.053 	& 1.15 & 2.57 & 4.00 & 9.97	\\	\hline
			VAE\cite{Neumeier.2021}			 & 253.736	& 4.21 & 4.38 & 3.45 & 7.36	\\	\hline
			CS-LSTM		 & 556.362	& 3.27*\textsuperscript{\rm 1}/2.88 & 5.71 & 8.65 & 16.93\\	\hline
			MHA-LSTM(+f) & 673.305 & 1.18*\textsuperscript{\rm 1}/2.58 & 5.44 & 13.10 & 27.45	\\	\hline
			Two-channel & 80.370 & 2.97 & 6.30 & 6.55 & 14.13	\\	\hline
			RA-GAT & 91.578 & 3.46 & 6.93 & 4.23*\textsuperscript{\rm 2}/7.05 & 15.49 \\ \hline
			\hline
		\end{tabular}%
		\caption{Trajectory prediction performance of different AI-models including state-of-the-art approaches. Where needed, the grid definitions of the baseline models were adopted, so that the same traffic scenario is provided to each model. \newline
			Results marked with *\textsuperscript{\rm 1} are taken from \cite{Messaoud.2021} and *\textsuperscript{\rm 2} from \cite{RAGAT.2021}, since their problem definition is equal to the one of this work. }
		\label{tab:comparison}
	\end{table*}
	\section{Experimental Evaluation}
	In order to evaluate the prediction performance of the GFTNN, it is compared to state-of-the-art approaches in the task of vehicle trajectory prediction. Thereby, prediction should only be based on traffic dynamics and no additional network to handle the infrastructural conditions is added. To yet allow a fair comparison with baseline models, datasets and models where infrastructural information is crucial for the prediction performance are excluded. The proposed GFTNN is hence benchmarked only upon highway datasets. 
	For training and testing the publicly available datasets highD \cite{highD} and NGSIM \cite{ngsim.2017} are used due to their extend of application-oriented scenarios. The datasets provide traffic observations of different German or US highways, respectively. The raw datasets, however, hold a naturally huge imbalance of scenarios since lane changes happen less often than lane keepings. When training on such a highly unbalanced data, a pretty high accuracy can be achieved by solely predicting the trajectory belonging to the majority class. Simultaneously, the model most probably fails to capture the minority class of lane change predictions. This problem was investigated in more depth by Ding \textit{et al.} \cite{RAGAT.2021}:
	As analyzed, about \SI{96.37}{\percent} of the scenarios contained in NGSIM dataset are keep lane scenarios. While the performance of their introduced trajectory prediction model on the overall dataset is superior to other baseline models, it performs poorly on the lane change scenarios. 
	In order to enable fair benchmarking, each dataset is pre-selected and pre-processed in a fashion, that the contained trajectory scenarios are balanced. Furthermore, the global description of the datasets is transformed into relative (target vehicle centered) descriptions. The resulting data format aligns with the problem definition explained beforehand. In total, $N_V = 9$ vehicles are considered. This definition allows the inclusion of the eight possible immediate positional neighbors and the target vehicle itself.
	 When a scenario holds too many vehicles, the participants with the smallest Euclidean distance to the target vehicle are selected. When a scenario lacks participants, "ghost vehicles" are inserted. These ghost vehicles are attributed with the same motion features as the target vehicle itself, so that no artificially simulated dynamic is added. 
	Each scenario consists of an observation period and a prediction period. The observation period is set to $\SI{3}{\second}$ so that $T_{\text{obs}}$ is adaptively parametrized by the dataset frequency ($fps_{\text{highD}} = \SI{25}{\hertz}$, $fps_{\text{NGSIM}} = \SI{10}{\hertz}$). The prediction horizon is set to $t_\text{pred} = \SI{5}{\second}$ and the number of prediction steps result from the respective dataset frequency.
	From the highD/NGSIM dataset 9000/1100 highway scenarios are extracted, that represent an equal distribution of the maneuvers \textit{keep lane}, \textit{lane change right} and \textit{lane change left}. The reduced amount of scenarios for the NGSIM results from limited lane change scenarios in the dataset itself.  
	\subsection{Implementation Details}
	With the scope of implementing a lightweight neural network, the MLP block is set to $N_x=1$. The used encoder architecture hence holds one element-wise multiplication layer and three MLP layers with the dimensions $T_{\text{obs}[0:p-1]}\times N_V-50-3$ for each feature $K$. In the following, different graph and hyperparamter settings for the GFTNN, which are included into evaluation, are explained. 
	
	\textbf{gftnn}: The spatio-temporal graph is an undirected and unweighted Cartesian product graph. The node feature dimension is set to $K=4$ so that all features of $\boldsymbol{\xi}_j$ are considered in the GFT. The complete spectrum is forwarded to the encoder, hence $p=T_{\text{obs}}$. 
	
	\textbf{gftnn-w}: The spatio-temporal graph is an undirected and \textit{weighted} Cartesian product graph. The weights are defined by the inverse Euclidean distance to the target vehicle. The node feature dimension is therefore reduced to $K=2$, only considering the velocities of $\boldsymbol{\xi}_j$ in the GFT. The complete spectrum is forwarded to the encoder, hence, $p=T_{\text{obs}}$. Since the graph weighting is adapted according to the scenario, the aggregation scheme is non-injective.
	
	\textbf{gftnn-rdcby5/15}: The spatio-temporal graph is an undirected and unweighted Cartesian product graph. The node feature dimension is set to $K=4$ so that all features of $\boldsymbol{\xi}_j$ are considered in the GFT. Only a reduced (rdc) fraction of the complete spectrum is forwarded to the encoder:\\ rdcby5 $\rightarrow$ $p=T_{\text{obs}}/5$, rdcby15 $\rightarrow$ $p=T_{\text{obs}}/15$. 
	
	The train-test split is set to 70-30. Optimization is done using Adam, set to a learning rate of $lr=1e-4$ within 30 (highD) or 120 (NGSIM) epochs, respectively. The loss function $\mathcal{L}$ is the mean squared prediction error (MSE)
	\begin{equation}
	\mathcal{L} = \text{MSE}(\bm{x},\bm{\hat{x}}) + \text{MSE}(\bm{y},\bm{\hat{y}}).
	\end{equation}
	
	\subsection{Evaluation}
	\begin{figure}
		\centering
		\vspace{5pt}
		\def\svgwidth{0.9\columnwidth} 
		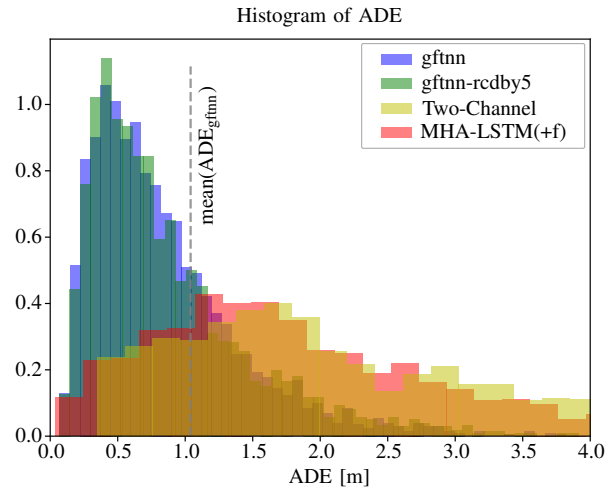
		\caption{Average displacement error histogram on highD trajectory predictions. The GFTNN variants show an exponential decrease for error values greater than $\sim\SI{0.4}{\meter}$. In contrast, the baseline models indicate a more frequent occurrence of greater prediction errors.}
		\label{fig:errorDist}
	\end{figure}
	The GFTNN is compared with state-of-the-art baselines including conventional recurrent models (CS-LSTM \cite{Deo.2018}, MHA-LSTM(+f) \cite{Messaoud.2021}) and novel graph-based approaches (Two-Channel GNN \cite{Mo.08.07.2021}, RA-GAT \cite{RAGAT.2021}). For evaluation the widely used metrics \textit{Average Displacment Error (ADE)}
	\begin{equation}
	\text{ADE} = \sqrt{\frac{ \sum_{i=1}^{N} \frac{1}{T_\text{pred}}\left(
			(\bm{x}^{(i)} - \hat{\bm{x}}^{(i)})^2 + (\bm{y}^{(i)} - \hat{\bm{y}}^{(i)})^2\right)}{N}}
	\end{equation}
	and \textit{Final Displacement Error (FDE)}
	\begin{equation}
	\text{FDE} = \frac{ \sum_{i=1}^{N} \sqrt{(\bm{x}_{T_{\text{pred}}}^{(i)} - \hat{\bm{x}}^{(i)}_{T_{\text{pred}}})^2 + (\bm{y}^{(i)}_{T_{\text{pred}}} - \hat{\bm{y}}^{(i)}_{T_{\text{pred}}})^2}}{N}
	\end{equation}
	are applied. As shown in Table \ref{tab:comparison}, GFTNN achieves the best results on both metrics and datasets.
	More specifically, applied on the highD dataset it achieves a performance improvement of $\sim\SI{13}{\percent}$ on ADE with respect to the results  provided by literature of the best performing model. 
	Evaluation based on the balanced dataset used for this work emphasizes the prediction performance of GFTNN:the error metrics ADE and FDE indicate a performance improvement of $\sim\SI{60}{\percent}$ and $\sim\SI{45}{\percent}$, respectively, with respect to the best performing baseline model. Considering the results of NGSIM the performance of all models decreases.
	Yet, the prediction performance of the GFTNN without parameter reduction is still superior. On this dataset, the best performing baseline model is the LSTM-based VAE \cite{Neumeier.2021}, which the vanilla gftnn outperforms by $\sim\SI{12}{\percent}$. Since average values do not depict the error variance, the prediction performance is additionally evaluated through the error distribution. In Fig.~\ref{fig:errorDist} the distribution of the ADEs resulting from the best performing models on highD is illustrated. Both GFTNN implementations represent distributions that have their peaks at an ADE of $\SI{0.4}{\meter}$ and exponentially decrease as the prediction error increases. The compared models Two-Channel and MHA-LSTM(+f) show a suboptimal error distributions that indicate a worse prediction accuracy in general. The distribution peaks of both models ($\SI{1.7}{m}/\SI{1.2}{m}$) indicate a greater universal prediction error. 
	Evaluation of the different hyperparameter settings for the GFTNN shows, that the best performing model is the gftnn setup. For the highD dataset, however, performance does not drastically decrease even when the model complexity is reduced as in gftnn-rdcby15. This equals a parameter reduction to $\SI{7}{\percent}$ of the original number of parameters. Applied on the NGSIM dataset, the reduction of parameters affects the prediction performance substantially. Note here, that the frequency provided by the NGSIM dataset (\SI{10}{\hertz}) is lower than the one of the highD (\SI{25}{\hertz}). Consequently, also the embedded temporal spectra using the NGSIM dataset provide a reduced frequency range. By neglecting the high frequencies of the NGSIM spectra essential dynamics information for the motion prediction is removed. This character indicates that when data frequency is low, high frequency neglection leads to inaccurate predictions.
	Furthermore, prediction performance decreases when using a non-injective neighborhood aggregation scheme.
	The gftnn-w adapts its spectral transformation for each scenario by weighting the graph edges through distance information. Even though this approach is provided the same information as the non-weighted GFTNNs, prediction accuracy generally decreases.
	\subsection{Interpretation of Prediction Performance}
	The GFT is a global aggregation scheme, which comprises the complete scenario graph instead of several local $k$-hop neighborhoods. The global characteristic is important for the understanding of the cooperative context of the scenario and ensures the sequence consideration without recurrent layers. By this transformation, the multidimensional context of the graph signal is retained. Since the Fourier transformation is an integral transform, the resulting spectrum holds the information of the complete time series. The spectral feature space indicates how smooth the prediction vehicle's movement is within the cooperative spatio-temporal context. By explicitly providing the network with relational information through the graph structure, a lot of additional knowledge is inserted that fortifies the learning process.
	The resulting relation-based spectral scenario representation is a beneficial mapping, which facilitates the neural network to capture the scenario context and extract the key information for trajectory predictions. 
	
	\section{Conclusion}
	In this work, the GFTNN is proposed which combines the advantages of using graph structures and the expressive power of conventional FNNs. By representing a traffic scenario through a spatio-temporal graph and applying a multidimensional GFT, an efficient time-space-relation representation of the scenario is obtained.
	Even though the overall GFTNN architecture is kept computationally simple, it outperforms the baseline models in the task of vehicle trajectory prediction on highways. Quantitatively, the vanilla GFTNN setup surpasses the best performing state-of-the-art method by $\sim \SI{13}{\percent}$ on the highD dataset and $\sim \SI{12}{\percent}$ on the NGSIM.
	The architecture enables to create a lightweight GFTNN version with highly reduced complexity and parameters while still achieving competitive performance to state-of-the-art networks.
	In future work, the intuition behind the graph properties and the usage of non-static graphs will be addressed. Furthermore, the GFTNN will be extended to include infrastructural scenario information and compared to state-of-the-art models that consider infrastructure, like the SpecTGNN \cite{Cao.5302021652021}.

	\section*{A. Appendix: GFT on static graphs} \label{appendix}
	The 2D GFT is represented as a chain of matrix multiplications. By using $N_1 \times N_2$ matrices $\bm{F}_{i_1,i_2} = f(i_1, i_2)$ and $\hat{\bm{F}}_{k_1,k_2} = \hat{f}(\lambda^{(1)}_{k_1}, \lambda^{(2)}_{k_2})$, the 2D GFT applied to the signal $f$ is expressed as
	\begin{equation}
	\mathcal{F}(\bm{F}) = \bm{\hat{F}} = 
	\bm{U}_1^*
	\bm{F}
	\overline{\bm{U}}_2
	\end{equation}
	where $\bm{U}_n$ is an $N_n \times N_n$ unitary matrix with ($i, k$)-th element $u_k^{(n)}(i)$ for $n=1,2$. Let $\bm{U}_n^*$ be the Hermitian transpose and $\overline{\bm{U}}_n$ the element-wise complex conjugate matrix.
	The eigenvector matrices $\bm{U}_n \in \mathbb{R}^{N_n \times N_n}$ result from the constant Laplacian matrices which are real, symmetric and positive-semidefinite.
	\subsection*{Proof of injective characteristic} A function is injective if $\forall a,b \in \mathcal{X}, \mathcal{F}(a)= \mathcal{F}(b) \rightarrow a = b$. Assume $\mathcal{F}(\bm{F}_A) =
	\bm{U}_1^*	\bm{F}_A \overline{\bm{U}}_2$ and $\mathcal{F}(\bm{{F}}_B) = \bm{U}_1^*	\bm{F}_B \overline{\bm{U}}_2$ where $\mathcal{F}: \mathbb{R}^{N_1 \times N_2} \rightarrow \mathbb{R}^{N_1 \times N_2}$.
	\begin{align}
	\mathcal{F}(\bm{F}_A) &= \mathcal{F}(\bm{F}_B)\\
	\bm{U}_1^*	\bm{F}_A \overline{\bm{U}}_2 &= \bm{U}_1^*	\bm{F}_B \overline{\bm{U}}_2\\
	\bm{F}_A &= \bm{F}_B
	\end{align}
{\small
	\bibliographystyle{ieee_fullname}
	\bibliography{ref/ref_gilbert.bib}
}

\end{document}